\documentclass[10pt,twocolumn,letterpaper]{article}

\usepackage{cvpr}              

\usepackage{graphicx}
\usepackage{amsmath}
\usepackage{amssymb}
\usepackage{booktabs}
\usepackage{algorithm}
\usepackage{algorithmic}

\usepackage[pagebackref,breaklinks,colorlinks]{hyperref}

\usepackage[capitalize]{cleveref}
\crefname{section}{Sec.}{Secs.}
\Crefname{section}{Section}{Sections}
\Crefname{table}{Table}{Tables}
\crefname{table}{Tab.}{Tabs.}

\def\cvprPaperID{*****} 
\def\confName{CVPR}
\def\confYear{2022}

\begin{document}

\title{1st Place Solutions for CVPR 2022 CLVision Challenge Track 2 and Track 3  \textit{nVFNet-RDC: Replay and Non-Local Distillation Collaboration for Continual Object Detection}}

\author{
\textit{Track2}:
Jinxiang Lai, 
Wenlong Liu, \\
Wenlong Wu, 
CongChong Nie,
ShuangShuang Guo,
Xing Gong,
Jun Liu \\
\textit{Track3}:
Jinxiang Lai, 
Wenlong Liu,
Tao Wu, 
Yi Zeng, 
Jialin Li, 
Yong Liu,
Jun Liu \\
Tencent Youtu Lab, Shenzhen, China \\
{\tt\small \{jinxianglai, juliusliu\}@tencent.com}
}
\maketitle

\begin{abstract}
Continual Learning (CL) focuses on developing algorithms with the ability to adapt to new environments and learn new skills.
This very challenging task has generated a lot of interest in recent years, with new solutions appearing rapidly.
In this paper, we propose a nVFNet-RDC approach for continual object detection.
Our nVFNet-RDC consists of teacher-student models, and adopts replay and feature distillation strategies.
As the 1st place solutions, we achieve 55.94\% and 54.65\% average mAP on the 3rd CLVision Challenge Track 2 and Track 3, respectively.
\end{abstract}

\vspace{-2mm}
\section{Introduction}
\label{sec:intro}
In this report, we introduce the 1st Place Solutions for the 3rd CLVision Challenge \cite{CLVISION2022} Track 2 \cite{track2} and Track 3 \cite{track3}, which are continual category-level and instance-level object detection, respectively.
Our solution is called \textit{nVFNet-RDC: Replay and Non-Local Distillation Collaboration for Continual Object Detection}.
In the following, we will first give a brief introduction for EgoObjects Dataset and Benchmark Setting.
Then, we will introduce our Solutions for CLVision Track 2 and Track 3, Experiments and Failed Attempts.

\begin{figure}[!t]
\centering
\includegraphics[width=0.99\linewidth]{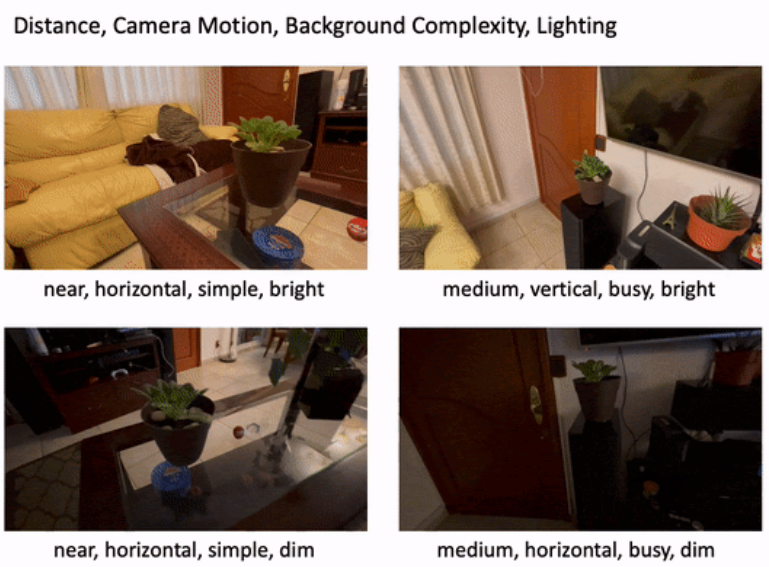}
\vspace{-2mm}
\caption{The EgoObjects Dataset.}
\label{fig:ego}
\vspace{-4mm}
\end{figure}

\begin{figure}[ht]
\centering
\includegraphics[width=0.99\linewidth]{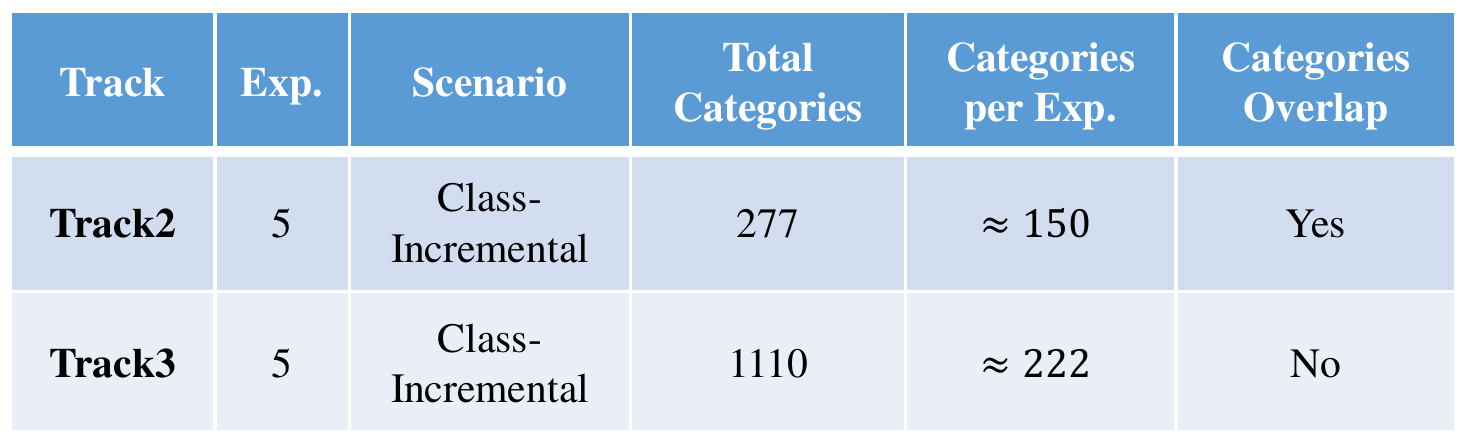}
\vspace{-2mm}
\caption{The benchmark setting of Track 2 and Track 3.}
\label{fig:benchmark}
\vspace{-4mm}
\end{figure}

\begin{figure*}[!t]
\centering
\includegraphics[width=0.99\linewidth]{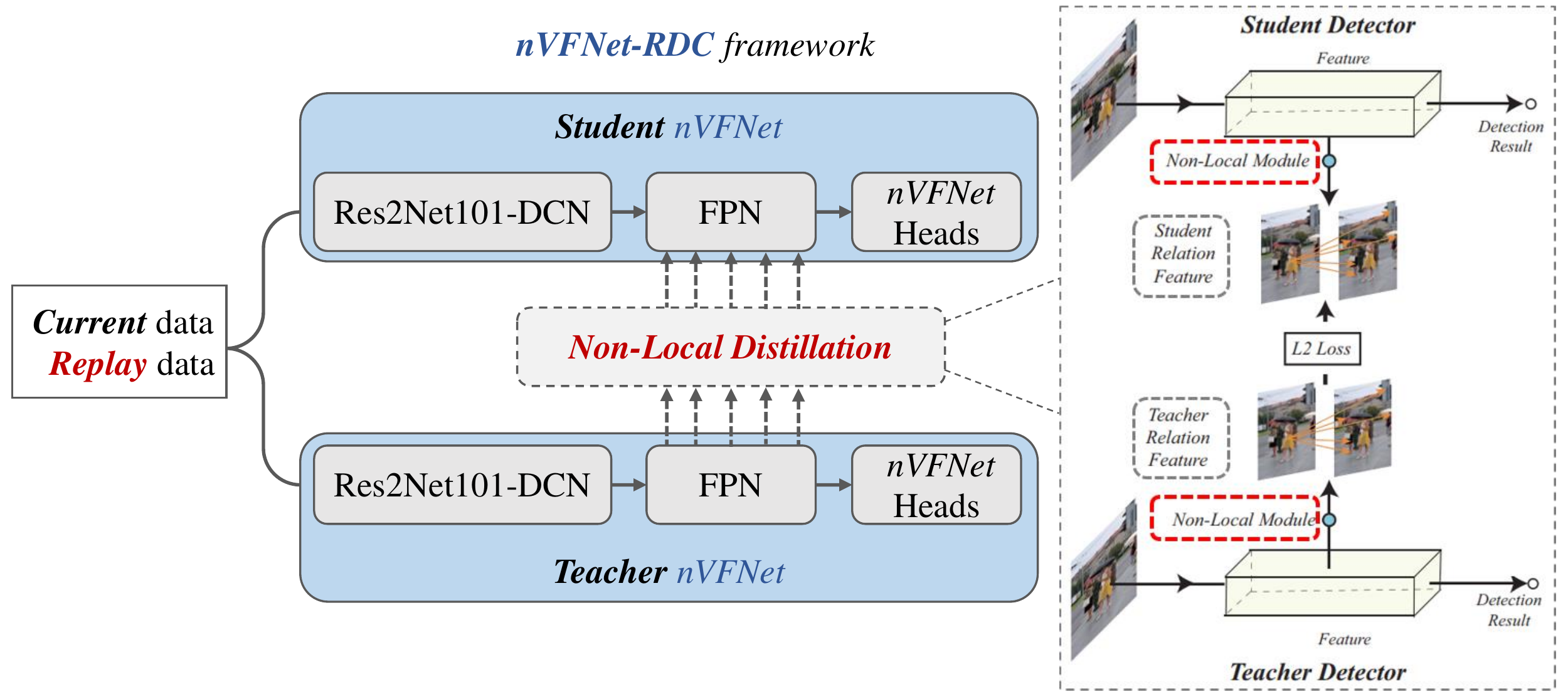}
\vspace{-2mm}
\caption{The framework of our nVFNet-RDC. Upon the proposed nVFNet detector, the continual learning approaches containing Replay and Non-Local Distillation are integrated.}
\label{fig:framewrok}
\vspace{-4mm}
\end{figure*}

\subsection{The EgoObjects Dataset}
\label{subsec:dataset}
The challenge is supported by the EgoObjects dataset provided by Meta, a massive-scale egocentric dataset for objects, which contains around 70 thousand images for training and 10 thousand images for testing. 
Critical features of this dataset are as follows:
\begin{itemize}
\vspace{-2mm}
\item
As shown in Fig.~\ref{fig:ego}, videos feature a great variety of lighting conditions, scale, camera motion, and background complexity.
\vspace{-2mm}
\item
Video frames have been sampled and annotated with rich ground truth, including object category, object instance ID, 2D bounding box. 
\vspace{-2mm}
\item
Each video depicts one main object. In the same video, the surrounding objects are also annotated.
\vspace{-2mm}
\item
The scene complexity (amount of objects, occlusions, etc) is less than the one found in COCO, but the image quality is more varied.
\end{itemize}

\subsection{Benchmark Setting}
\label{subsec:challenge}
As shown in Fig.~\ref{fig:benchmark}, both Track 2 and Track 3 are belonging to Class-Incremental objection detection, and our solution is general for both of them.

The Evaluation Metric is Average mAP.
Specifically, for each experience testing on whole test set, the mAP is calculated. Then, all experience results are averaged to get the final score.
The train and test sets contains 73,905 and 10,713 images from 1110 scenes, respectively.
Incremental experiences will carry short videos of common household or workplace objects:
\begin{itemize}
\vspace{-2mm}
\item
For Track 2, objects will be depicted in common household and workplace environments, with each image depicting more than one object. The goal is to predict the bounding box and label of the depicted objects.
\vspace{-4mm}
\item
For Track 3, differently from its category-level counterpart, the goal for this track is to predict the object labels at the instance level. Each video will feature a single \textit{reference} object (possibly surrounded by other unrelated objects). The goal is to predict the position and instance label of that reference object.
\end{itemize}

\section{Methodology}
\label{sec:method}
\subsection{Overall nVFNet-RDC Framework}
Our solution for CLVision Track 2 and Track 3 is called nVFNet-RDC.
As shown in Fig.~\ref{fig:framewrok}, the nVFNet-RDC framework consists of teacher-student models, and adopts replay and feature distillation strategies.
We proposed a nVFNet Detector, which consists of VFNet and Non-Local Dense Classifier.
We propose a Continual Learning approach named RDC, which consists of Replay and Non-Local Distillation. 
Other Components includes: Res2Net backbone pre-trained on COCO, and PhotoMetricDistortion data augmentation.

As illustrated in Fig.~\ref{fig:framewrok}, with the replay strategy, the input data contains current data and replay data.
There is a teacher model trained from previous task, which consists of Res2Net backbone, FPN neck and detector heads.
And the student model needed to be trained for current task, which has the same structure with the teacher model.
Along with the conventional supervised loss, we further apply Non-Local Distillation for the features generated by FPN.

\begin{figure*}[!t]
\centering
\includegraphics[width=0.99\linewidth]{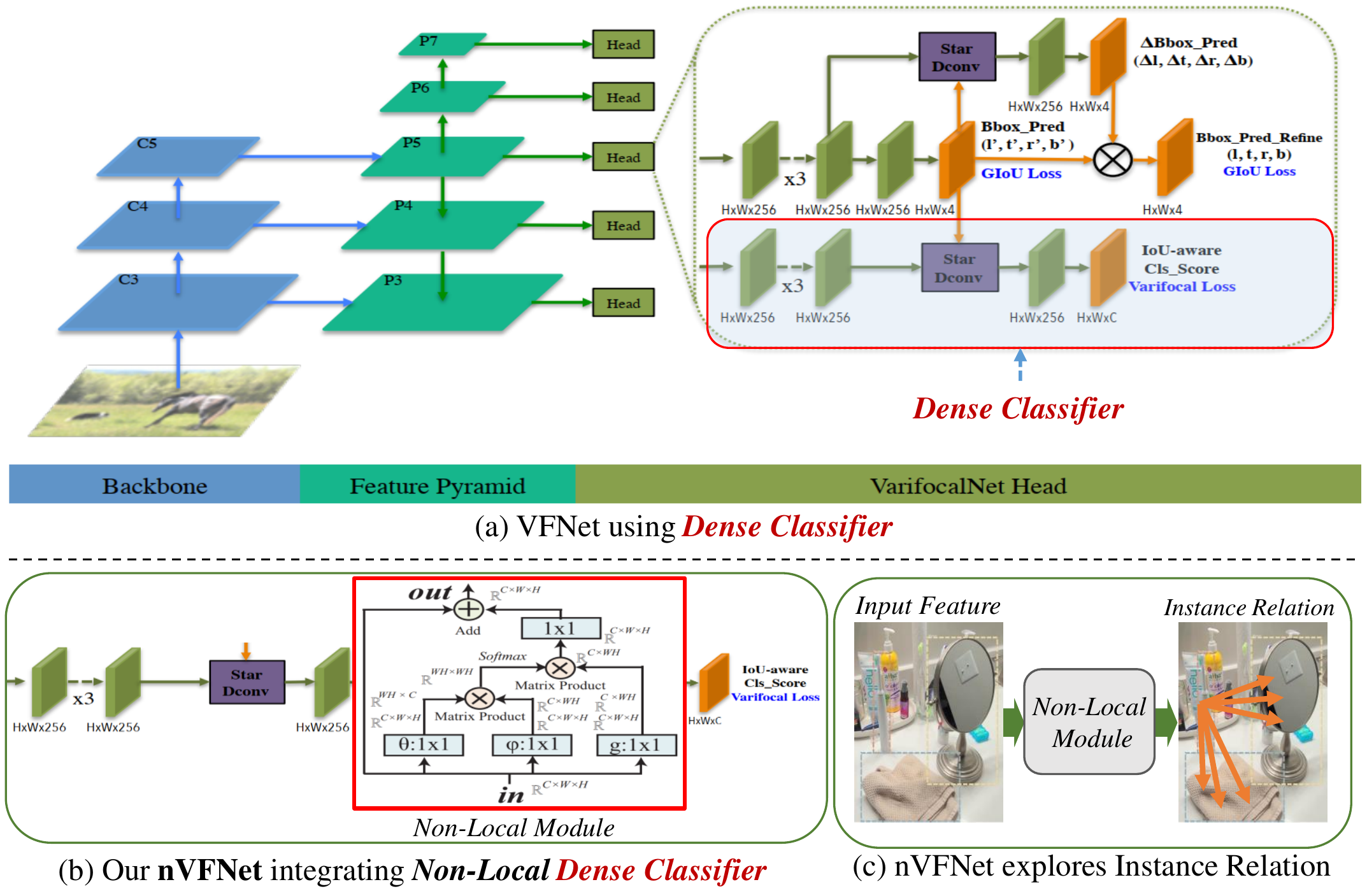}
\vspace{-2mm}
\caption{(a) The VFNet detector using Dense Classifier. (b) The proposed nVFNet detector integrating Non-Local Dense Classifier. (c) Our nVFNet explores the relations between instances.}
\label{fig:nvfnet}
\vspace{-4mm}
\end{figure*}

\subsection{nVFNet}
As illustrated in Fig.~\ref{fig:nvfnet}(a), the one stage VFNet \cite{zhang2021varifocalnet} is chosen as the base detector. 
The VFNet detector uses a dense classifier, which makes classification for each pixel of embedding features.
As illustrated in Fig.~\ref{fig:nvfnet}(b), we propose a nVFNet Detector via modifying the dense classifier of VFNet into a Non-Local Dense Classifier.
Specifically, a Non-Local Module \cite{wang2018non} is inserted into the Dense Classifier.
Given input feature to Non-Local Module, it models the relationship between instances, which is able to improve the accuracy of pixel-wise classification.

\begin{figure*}[!t]
\centering
\includegraphics[width=0.8\linewidth]{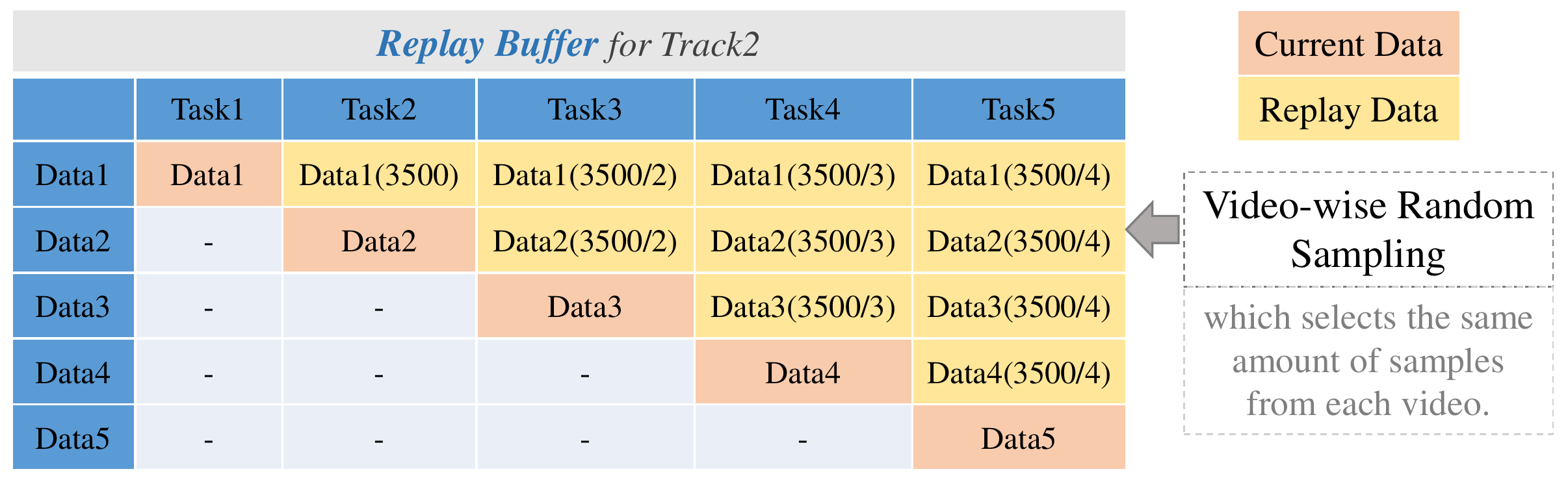}
\vspace{-2mm}
\caption{Our replay strategy.}
\label{fig:replay}
\vspace{-4mm}
\end{figure*}

\subsection{Replay Strategy}
Replay Rules are as follows: The replay buffer contains up to {3500 for Track 2, 5000 for Track 3} training samples. The buffer must be initially empty and can be populated using data from the current experience only. This means that the replay buffer cannot be populated beforehand, nor it can be filled with data from future experiences. The only samples from past experiences must be the ones chosen before terminating the training on that experience.

Fig.~\ref{fig:replay} shows our Replay Buffer. Task1 to Task5 are the 5 incremental tasks. Data1 to Data5 are the train sets for the corresponding tasks. The training data for each task is shown in each column, which contains current data and replay data. Best view in color, orange is current data, yellow is replay data. As we can see the replay data for each task, the same amount of data is sampled from the previous tasks.
We use Video-wise Random Sampling to obtain the replay data, which selects the same amount of samples from each video.

\subsection{Non-Local Distillation}
Inspired by \cite{zhang2020improve}, which improves object detection with feature-based knowledge distillation, we adopts feature-based Non-Local Distillation to handle the problem of catastrophic forgetting in continual learning.
As shown in Fig.~\ref{fig:framewrok}, a Non-Local Module \cite{wang2018non} is utilized to model the relationship between instances for the input features. Therefore, it distillates the relation features. Comparing to the directly feature distillation, the Non-Local Distillation is a weak feature distillation, which enables the model to learn a more discriminative embedding features for current task and also alleviates the forgetting problem.

\renewcommand{\tabcolsep}{3.5pt}
\begin{table*}[ht]
\caption{Experiments for Track 2. (1) Replay improves 14.1\%. (2) COCO pre-trained improves 4.08\%. 
(3) Non-Local Distillation improves 1.82\%. (4) Non-Local Dense Classifier improves 0.56\%. 
(5) PMD (PhotoMetricDistortion) improves 0.02\%. (6) Res2Net101-DCN is 2.49\% higher than ResNet50.
} 
\centering
\vspace{-0.3cm}
\begin{tabular}{c}
\begin{minipage}{2.3\columnwidth}
{\includegraphics[width=0.9\linewidth]{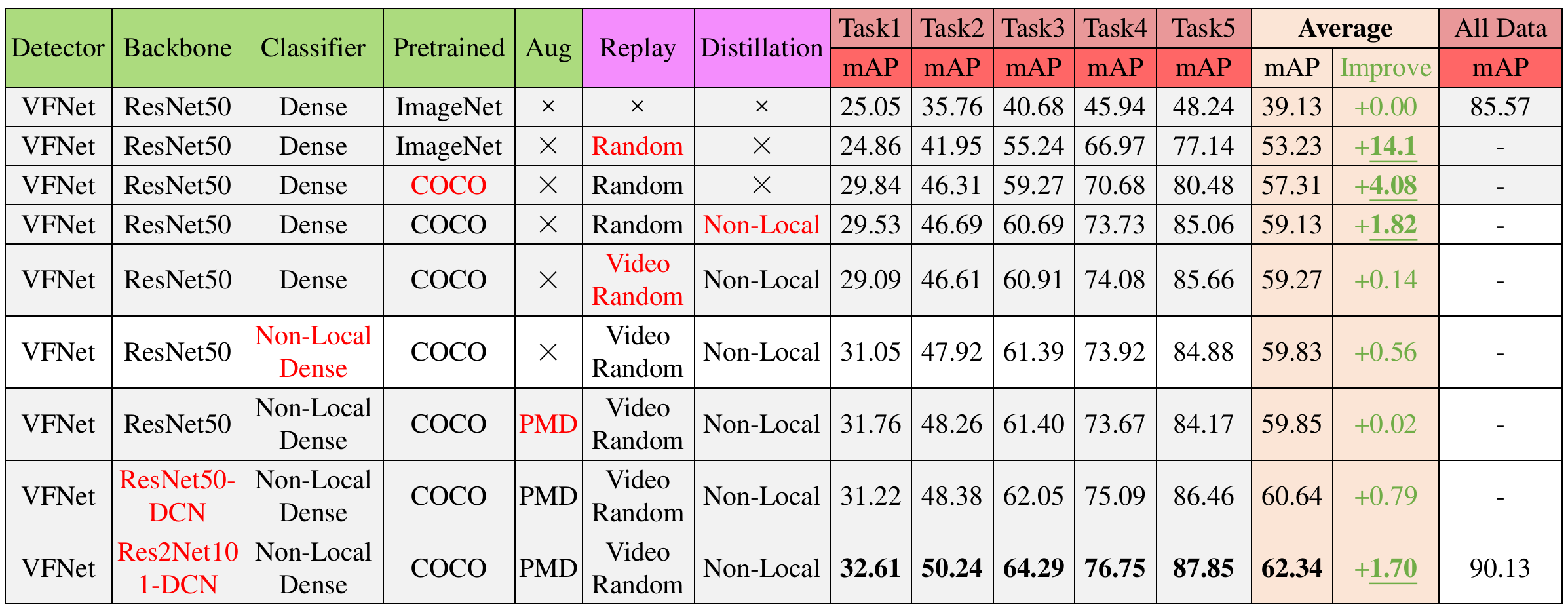}}  
\end{minipage}
\end{tabular}
\vspace{-0.4cm}
\label{table:track2}
\end{table*}

\renewcommand{\tabcolsep}{3.5pt}
\begin{table*}[ht]
\caption{Experiments for Track3. 
Replay and Non-Local Distillation achieves 8.85\% improvements.
} 
\centering
\vspace{-0.4cm}
\begin{tabular}{c}
\begin{minipage}[c]{2.3\columnwidth}
\includegraphics[width=0.9\linewidth]{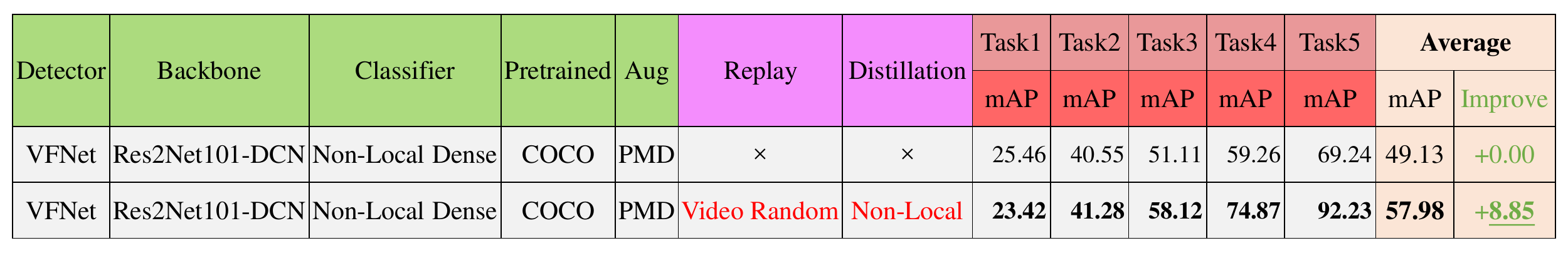}
\end{minipage}
\end{tabular}
\vspace{-0.4cm}
\label{table:track3}
\end{table*}

\section{Experiments}
\label{sec:experiments}
\subsection{Experiments for Track 2}
Tab.~\ref{table:track2} shows the improvements achieved by different components for Track 2. 
Best view in color, Green are the detector components, purple are continual learning strategies, the right sight are the results and we mainly focus on the Average results.
It indicates that:
\begin{itemize}
\vspace{-2mm}
\item
The first row is the baseline without any continual learning strategies.
\vspace{-2mm}
\item
The second row shows Replay strategy improves around 14\% than the baseline.
\vspace{-2mm}
\item
COCO pre-trained improves around 4\% than using ImageNet pre-trained.
\vspace{-2mm}
\item
Non-Local Distillation improves near 2\%.
\vspace{-2mm}
\item
Non-Local Dense Classifier improves around 0.5\%.
\vspace{-2mm}
\item
PhotoMetricDistortion data augmentation improves a little bit.
\vspace{-2mm}
\item
The deformable Res2Net backbone is around 2.5\% higher than ResNet50.
\end{itemize}

\subsection{Experiments for Track 3}
Tab.~\ref{table:track2} shows the improvements achieved by our approaches for Track 3. 
The proposed Replay and Non-Local Distillation achieves around 9\% improvements than the strong baseline.

\subsection{Ablation Study for Track 2}
\subsubsection{Influence of Detectors and Backbones}
Tab.~\ref{table:ablation_t2} shows that: first, VFNet is the better detector than others, such as Faster RCNN, ATSS, FCOS and GFL. 
Second, The performances of ResNet50 and Swin-T are pretty much the same.

\vspace{-2mm}
\subsubsection{Influence of Distillations}
Tab.~\ref{table:ablation_t2_2} shows that: first, Non-Local Distillation is better than others, such as directly Feature Distillation and Feature Attention Distillation. 
Second, Logit Distillation is not helpful upon Feature Distillation .

\section{Failed Attempts}
Tab.~\ref{table:fail} shows some Failed Attempts for Track2:
\begin{itemize}
\vspace{-2mm}
\item
The popular DETR and Deformable DETR are time consuming.
\vspace{-2mm}
\item
Multi-scale Data Augmentation is Not Effective. And Mosaic causes Performance Drop Heavily.
\vspace{-2mm}
\item
The long tail strategy Class Balance is Not Effective.
\vspace{-2mm}
\item
Adding Scene or Video Classification branch causes Performance Drop Heavily.
\end{itemize}

\begin{table*}[h]
\caption{Influence of Different Detectors and Backbones for Track 2.
(1) VFNet is the better detector than others. (2) The performances of ResNet50 and Swin-T are pretty much the same. 
} 
\centering
\vspace{-0.2cm}
\begin{tabular}{c}
\begin{minipage}[c]{2.3\columnwidth}
\includegraphics[width=0.9\linewidth]{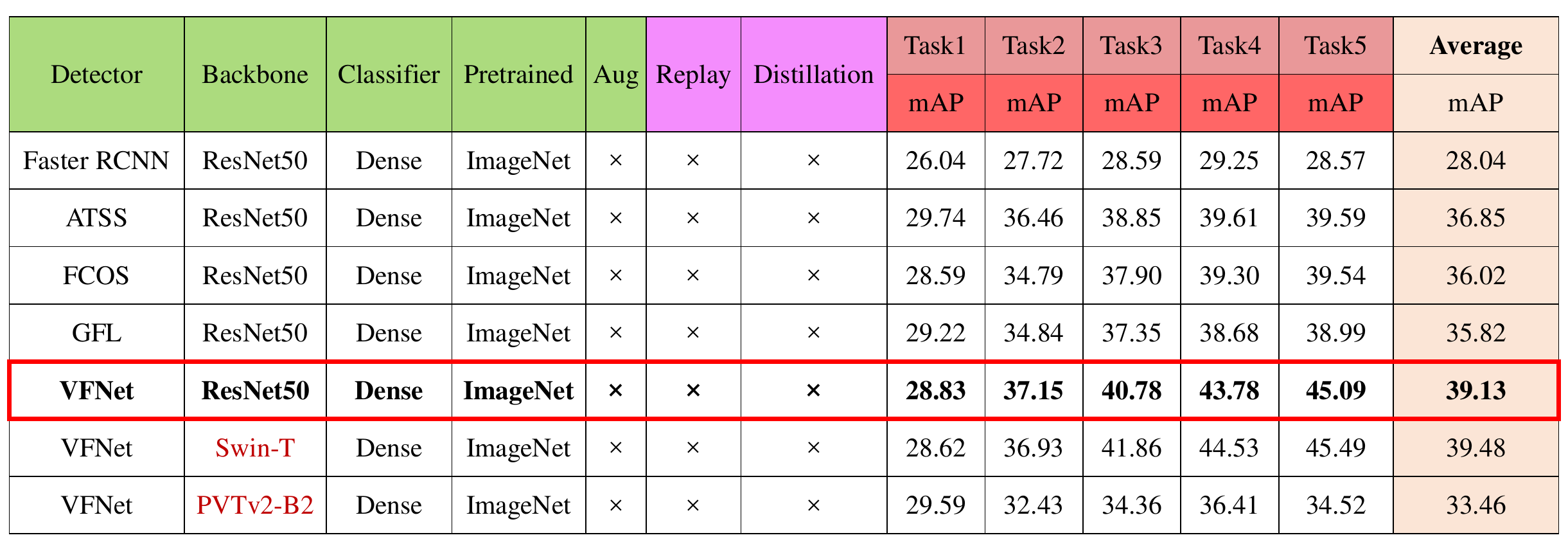}
\end{minipage}
\end{tabular}
\vspace{-0.4cm}
\label{table:ablation_t2}
\end{table*}

\begin{table*}[h]
\caption{Influence of Different Distillations for Track2.
(1) Non-Local Distillation is better than others. (2) Logit Distillation is not helpful upon Feature Distillation .
} 
\centering
\vspace{-0.2cm}
\begin{tabular}{c}
\begin{minipage}[c]{2.3\columnwidth}
\includegraphics[width=0.9\linewidth]{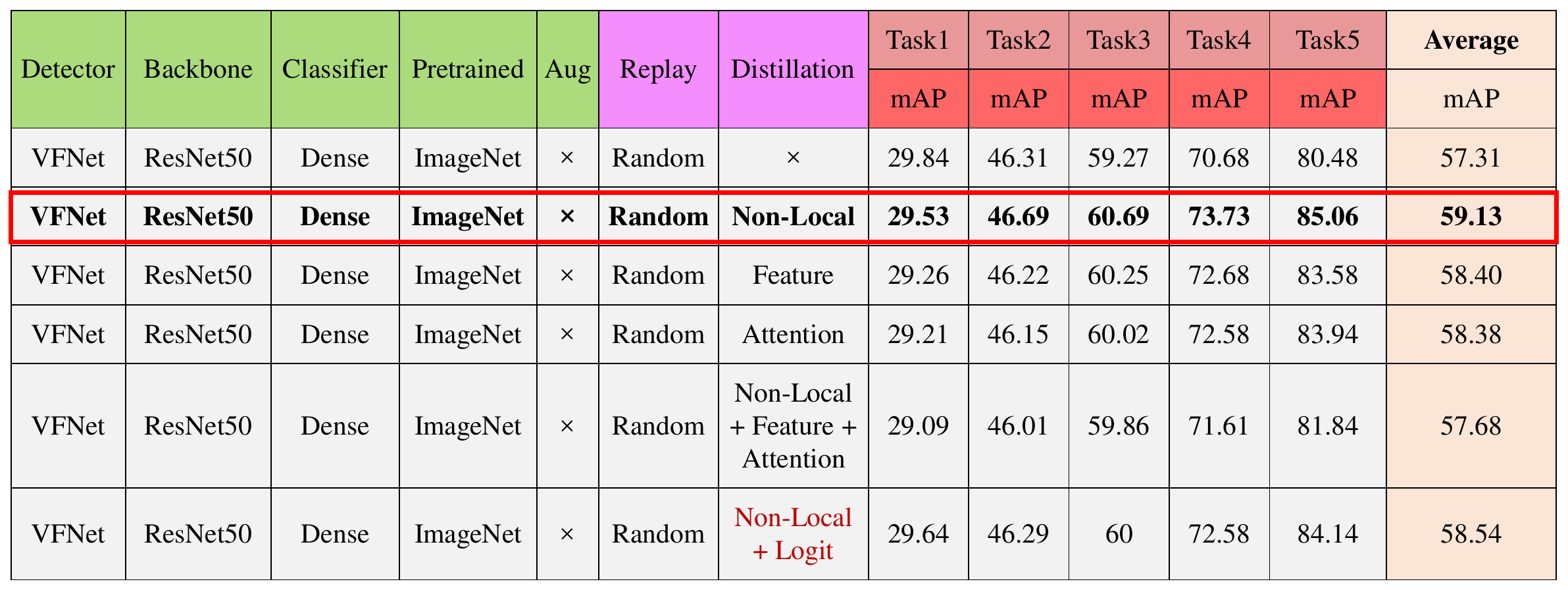}
\end{minipage}
\end{tabular}
\vspace{-0.4cm}
\label{table:ablation_t2_2}
\end{table*}

\begin{table*}[h]
\caption{Failed Attempts for Track 2.} 
\centering
\vspace{-0.4cm}
\begin{tabular}{c}
\begin{minipage}[c]{2.3\columnwidth}
\includegraphics[width=0.9\linewidth]{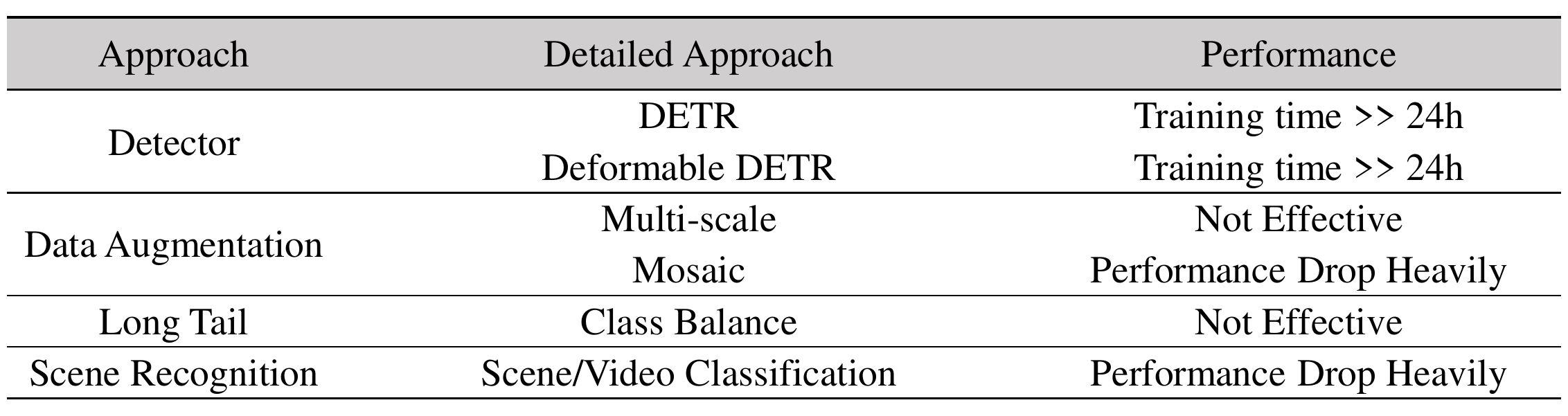}
\end{minipage}
\end{tabular}
\vspace{-0.4cm}
\label{table:fail}
\end{table*}

\clearpage
{\small
\bibliographystyle{ieee_fullname}
\bibliography{egbib}
}
\end{document}